%% file: main.tex
\documentclass{ifacconf}

\usepackage{color}
\usepackage[usenames,dvipsnames,table]{xcolor}
\usepackage{graphicx}      
\usepackage{natbib}        
\usepackage{kotex}
\usepackage{amsmath}

\usepackage{amssymb}  
\usepackage{subfigure}
\usepackage{subcaption}
\usepackage[normalem]{ulem}
\usepackage{siunitx}
\usepackage{url}

\input{symbol_definition}

\begin{document}
\begin{frontmatter}

\title{Demonstration of Space Robot Teleoperation over a Lossy and Delayed Network using ATMOS\thanksref{footnoteinfo}} 

\thanks[equal]{Inkyu Jang and Gregorio Marchesini contributed equally to this work.}
\thanks[corresponding]{Corresponding author. (e-mail: \url{janginkyu.larr@gmail.com)}}
\thanks[footnoteinfo]{This work was supported in part by the National Research Foundation of Korea
(NRF) grant funded by the Korea government (MSIT) (RS-2024-00436984), and in part by the Wallenberg AI, Autonomous Systems and Software Program (WASP) funded by the Knut and Alice Wallenberg (KAW) Foundation, the Swedish Research Council (VR), and Digital Futures.\\
(c) 2026 the authors. This work has been accepted to IFAC for publication under a Creative Commons License CC-BY-NC-ND.}


\author[snu]{Inkyu Jang\thanksref{equal}\thanksref{corresponding}}
\author[kth]{Gregorio Marchesini\thanksref{equal}}
\author[kth]{Nicola De Carli}
\author[snu]{Byeongjun Kim}
\author[snu]{Sunwoo Hwang}
\author[snu]{Dabin Kim}
\author[kth]{Elias Krantz}
\author[snu]{Youngkyoung Kong}
\author[fleetmq]{Frank J. Jiang}
\author[fleetmq]{Annika Wong}
\author[caltech]{Pedro Roque}
\author[kth]{Prasetyo W. L. Sanjaya}
\author[kth]{Nicola Bastianello}
\author[kth]{Mani H. Dhullipalla}
\author[kth]{Karl H. Johansson}
\author[snu]{Hyungbo Shim}
\author[kth]{Dimos V. Dimarogonas}
\author[snu]{H. Jin Kim}


\address[snu]{Seoul National University, 
   Seoul, Korea.}
\address[kth]{KTH Royal Institute of Technology, 
   Stockholm, Sweden.}
\address[fleetmq]{FleetMQ, 
   Stockholm, Sweden.}
\address[caltech]{California Institute of Technology, 
   Pasadena, CA, USA.}

\begin{abstract}                
\input{sections/abstract}
\end{abstract}


\end{frontmatter}
\begin{figure}
    \centering
    \includegraphics[width=0.85\linewidth]{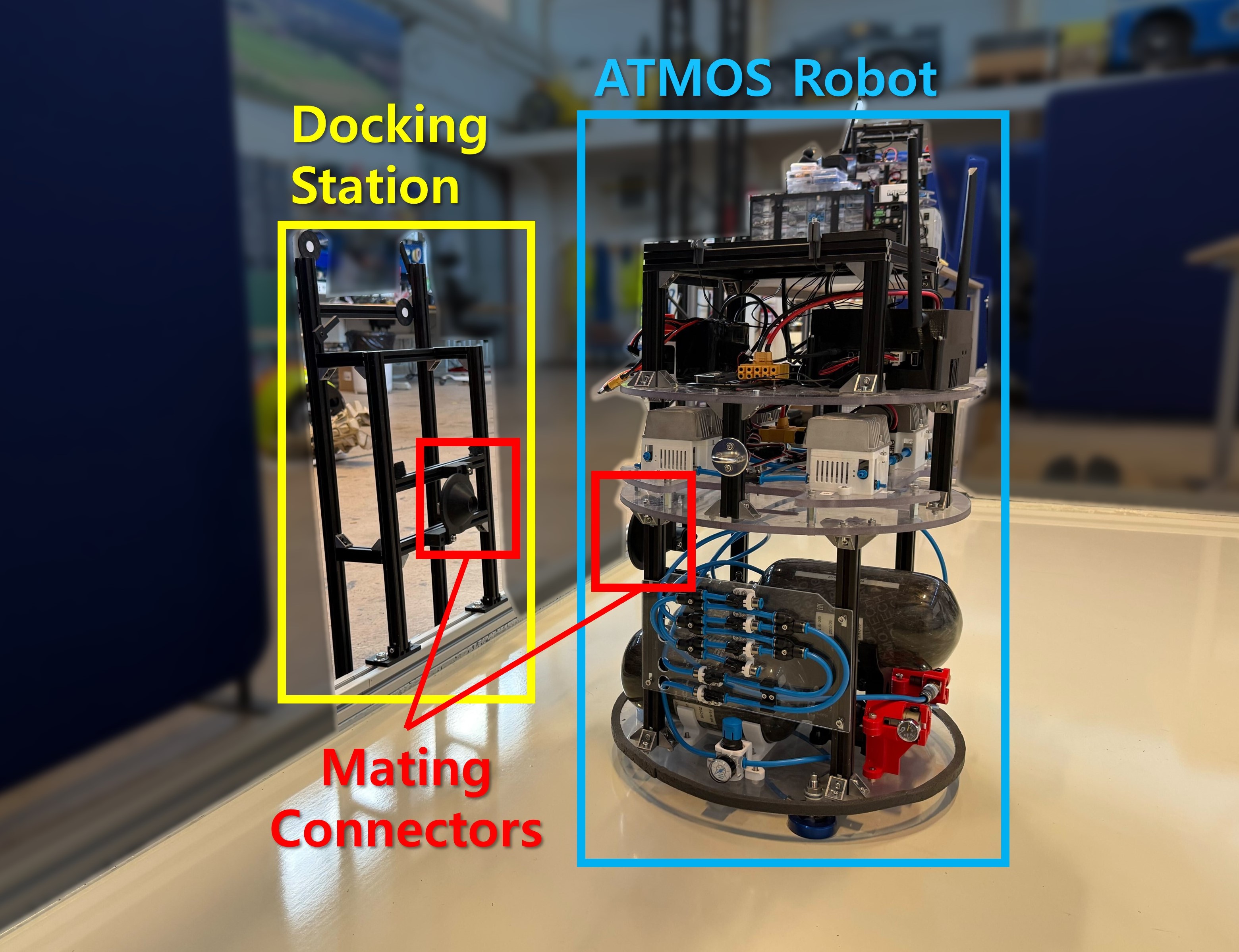}
    \caption{The robot hardware (ATMOS, \cite{roque2025towards}) considered in this demonstration.}
    \label{fig:thumbnail}
\end{figure}

\section{Introduction}
Recent technological advancements have fostered growing interest in the development of space missions for both terrestrial applications (e.g., climate monitoring, weather forecasting, gravitometry) and extraterrestrial exploration. As the demand for scientific throughput increases, remote-control architectures involving human-in-the-loop operators (\cite{teleoperations,teleoperations1}) as well as networked multi-agent solutions (\cite{liu2012network,orderreduction,wang2019distributed,zhang2024fixed}) have gained significant attention as reliable and cost-effective approaches to meet these mission requirements.
{These architectures depend on communication networks that may introduce delays, jitter, and packet loss, all of which can significantly degrade closed-loop performance.}


The reliability and predictability of these communication channels are therefore essential for safety-critical spacecraft applications (\cite{torgerson2024space}).
Designing control strategies that remain effective in the presence of such imperfections requires experimental platforms that can reproduce realistic communication conditions while enabling safe and repeatable testing on the ground.


In this demonstration paper, we showcase the Autonomy Testbed for Multi-purpose Orbiting Systems (ATMOS, \cite{roque2025towards}, Fig.~\ref{fig:thumbnail}) as a robotic platform for the rapid and cost-effective testing of delay-resilient spacecraft control strategies. ATMOS is a planar free-flying space robot that enables hardware-in-the-loop experiments for guidance and control algorithms under realistic conditions.

\begin{figure}
    \centering
    \includegraphics[width=\linewidth]{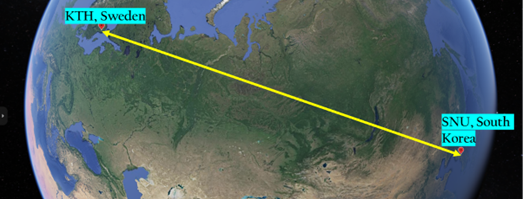}
    \caption{An illustration of the physical distance of 7450 \si{km} between the two labs in Stockholm, Sweden (the ATMOS robot) and Seoul, Korea (the controller).}
    \label{fig:globe}
\end{figure}

We develop a delay-aware Control Lyapunov Function (CLF) based controller (\cite{freeman1996control}) capable of tracking both fixed and time-varying reference trajectories despite stochastic, time-varying communication delays.
The main contribution of this work is that we present a transcontinental closed-loop control experiment between the space robotics laboratory at KTH Royal Institute of Technology (KTH) (Stockholm, Sweden) 
and Seoul National University (SNU) (Seoul, Korea).
In this setup, control inputs are computed in Seoul and transmitted to the ATMOS robot in Stockholm, while state measurements are streamed back to Seoul. The communications over this 7450 \si{km} distance as shown in Fig.~\ref{fig:globe} are made through the FleetMQ\footnote{FleetMQ - Seamless Vehicle Middleware. \url{https://www.fleetmq.com/}, accessed 2025-11-24.} middleware.
This introduces two-way latency profiles which affect the controller performance. Under the assumption that this latency comes from a slowly-varying distribution, we perform state prediction against delays based on previous observations of delay, and then apply the control in the direction to which the CLF value decays to achieve stability.

This demonstration highlights the capability of ATMOS to support the design and evaluation of delay-robust control architectures, providing an effective framework for validating teleoperation and autonomy concepts prior to on-orbit deployment.




\section{System Model}
\input{sections/system_model}

\section{Control Methodology}
\input{sections/control_methodology}

\subsection{State Prediction}

\input{sections/state_prediction}

\subsection{Reference Tracking Control}
\input{sections/delay_compensation_control}

\section{Simulation Experiments}

\begin{figure}
    \centering
    \includegraphics[width=0.9\linewidth]{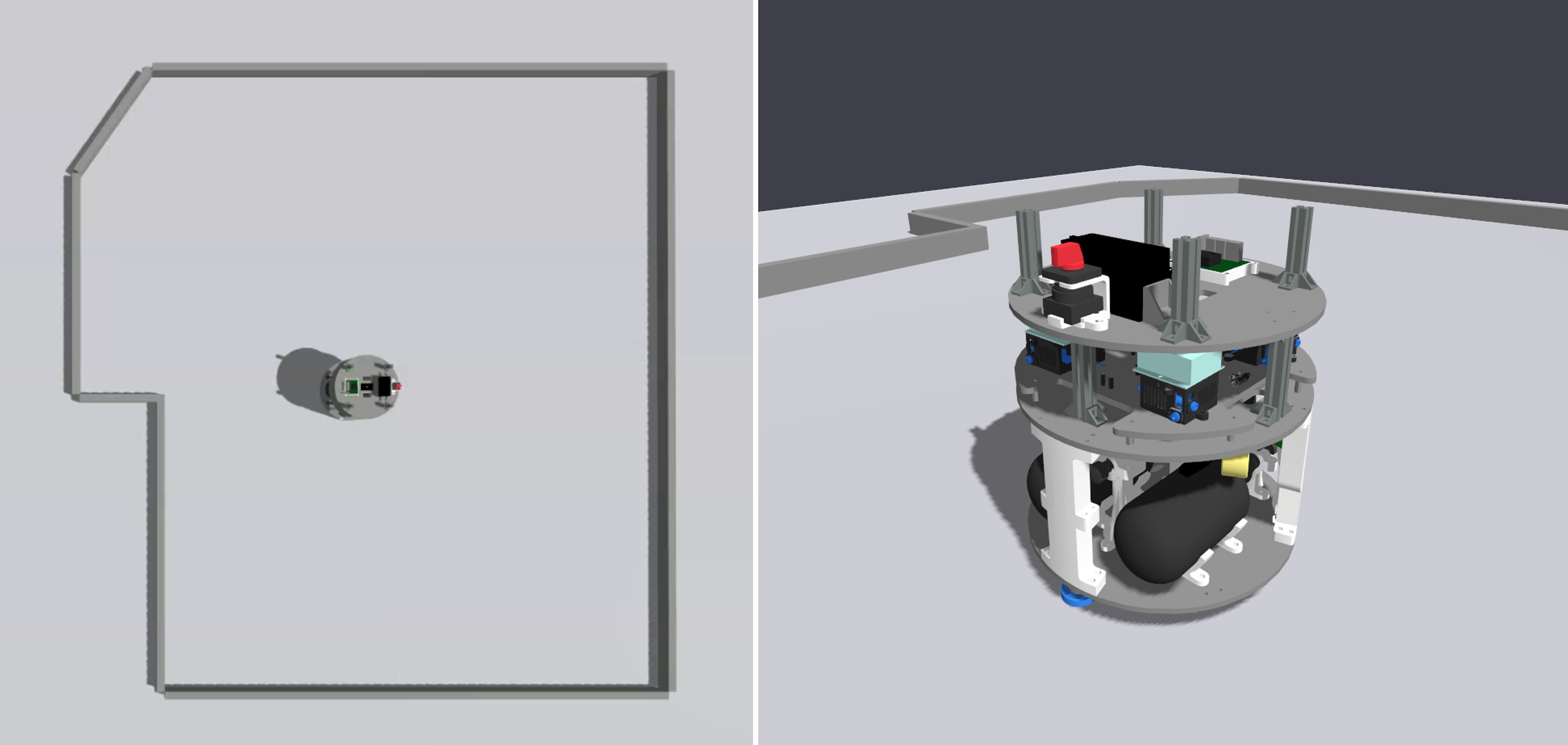}
    \caption{The Gazebo simulator environment used in the simulation experiments, replicating the real workspace of the KTH Space Robotics Laboratory (\cite{roque2025towards}). }
    \label{fig:gazebo}
\end{figure}

To showcase the validity of the proposed control method against delayed network, we first present Software-
In-The-Loop (SITL) simulation results using simulated stochastic delays. Compared to the hardware experiments, simulation experiments, where one is free to modify the delay characteristics, allow to analyze more on how the control performance varies with respect to different types of delays.
The simulation was done on a desktop computer equipped with a 2.40 GHz CPU and 32 GB RAM, running ROS 2, Gazebo simulator (Fig. \ref{fig:gazebo}) based on \cite{roque2025towards} and its accompanying software, and CVXOPT as the QP solver to solve (\ref{eq:clf controller}).

\begin{table}
\caption{Delay characteristics used in the \mbox{simulation experiments}}
\centering
\begin{tabular}{c|c|c|c}
\hline
 & $a$ [ms] & $b$ [ms] & Drop rate [\%] \\
\hline
No delay   & 0 & 0 & 0 \\
Light delay  & 100 & 200 & 20 \\
Moderate delay & 200 & 400 & 20 \\
Severe delay  & 300 & 600 & 20 \\
\hline
\end{tabular}
\label{tab:delay parameters}
\end{table}

\begin{table*}
\caption{Simulation result: Position and yaw RMSE under various delay levels}
\centering
\begin{tabular}{c||c|c|c|c||c|c|c|c}
\hline
& \multicolumn{4}{c||}{\textit{Stabilization} Task}
& \multicolumn{4}{c}{\textit{Tracking} Task} \\
\cline{2-9}
 & \multicolumn{2}{c|}{\textbf{Proposed}} 
 & \multicolumn{2}{c||}{Reference Only}
 & \multicolumn{2}{c|}{\textbf{Proposed}}
 & \multicolumn{2}{c}{Reference Only} \\
\cline{2-9}
 & \shortstack{Position [m]}
 & \shortstack{Yaw [$^\circ$]}
 & \shortstack{Position [m]}
 & \shortstack{Yaw [$^\circ$]}
 & \shortstack{Position [m]}
 & \shortstack{Yaw [$^\circ$]}
 & \shortstack{Position [m]}
 & \shortstack{Yaw [$^\circ$]} \\
\hline
No delay   & 0.022 & 0.536 & 0.020 & 0.430 & 0.069 & 0.767 & 0.081 & 8.511 \\
Light delay  & 0.022 & 1.515 & 0.025 & 1.714 & 0.060 & 1.496 & 0.118 & 15.456 \\
Moderate delay & 0.024 & 2.385 & 0.212 & 31.977 & 0.104 & 2.318 & 0.326 & 91.094 \\
Severe delay  & 0.032 & 4.421 & 0.618 & 76.208 & - & - & - & - \\
\hline
\end{tabular}
\label{tab:simulation rmse}
\end{table*}

\begin{figure*}[t]
    \centering

    \begin{minipage}{0.24\linewidth}
        \centering
        \includegraphics[width=\linewidth]{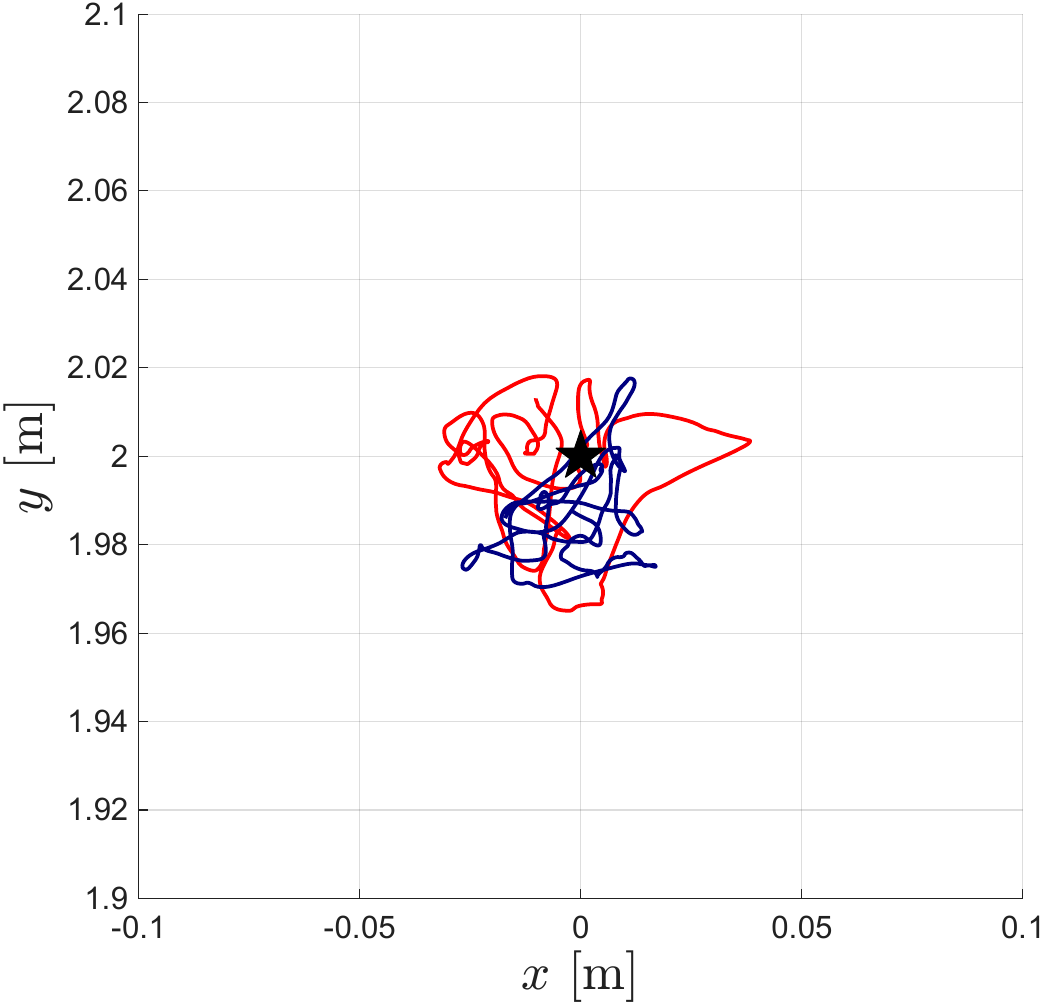}
        (a) No delay
    \end{minipage}
    \begin{minipage}{0.24\linewidth}
        \centering
        \includegraphics[width=\linewidth]{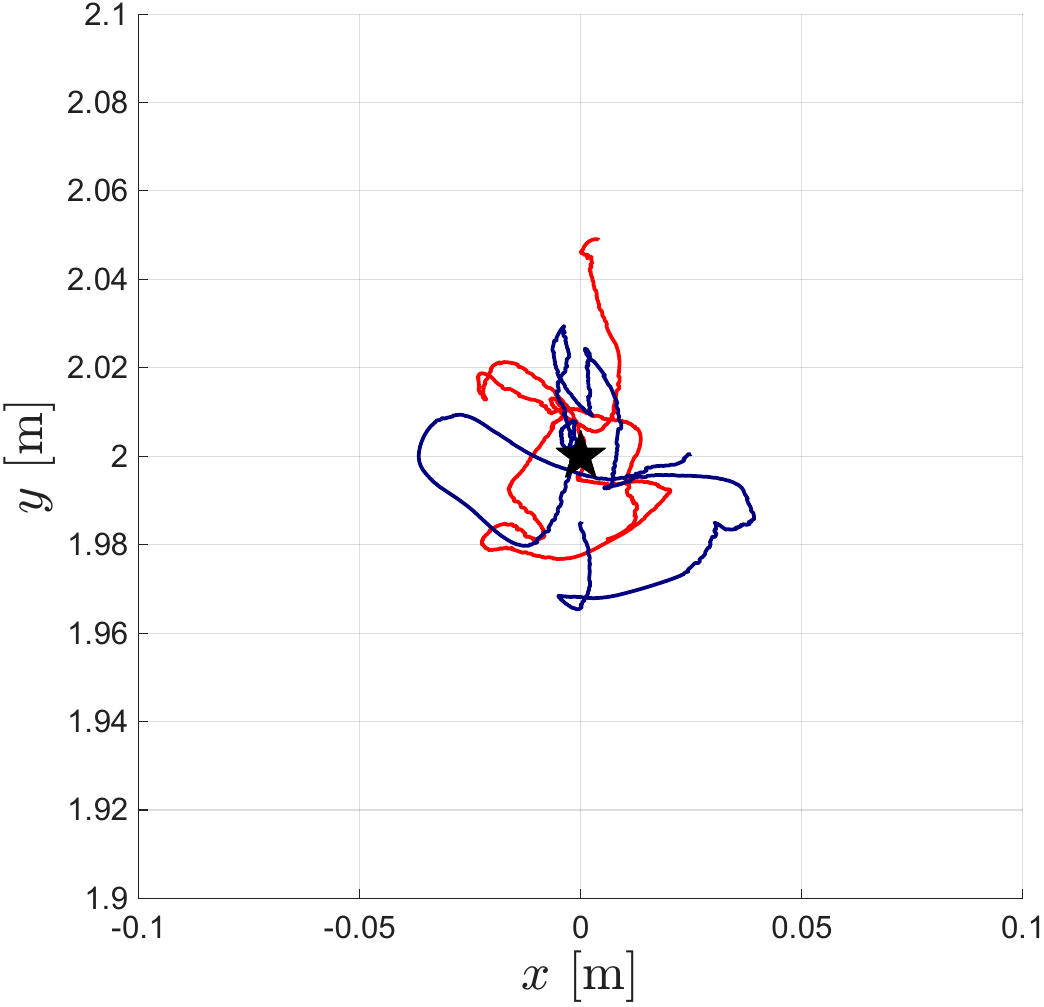}
        (b) Light delay
    \end{minipage}
    \begin{minipage}{0.24\linewidth}
        \centering
        \includegraphics[width=\linewidth]{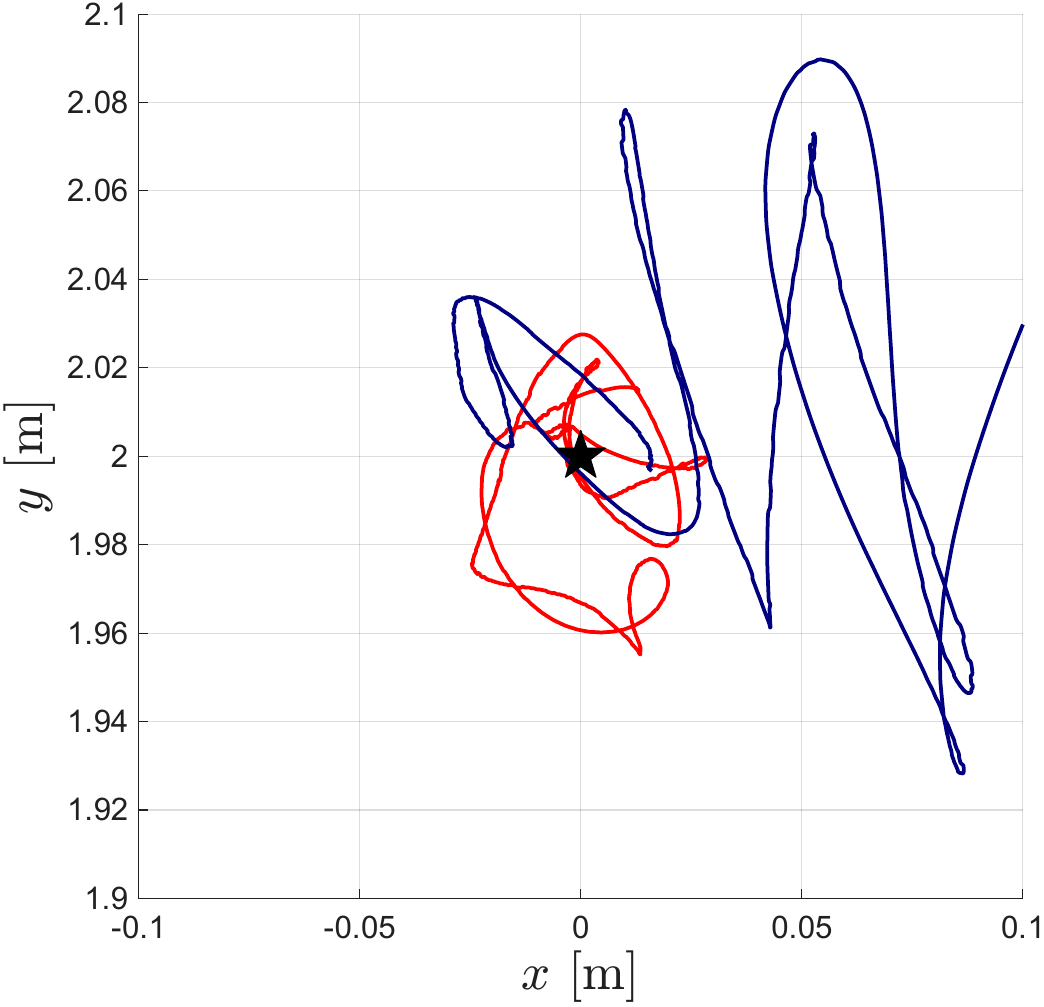}
        (c) Moderate delay
    \end{minipage}
    \begin{minipage}{0.24\linewidth}
        \centering
        \includegraphics[width=\linewidth]{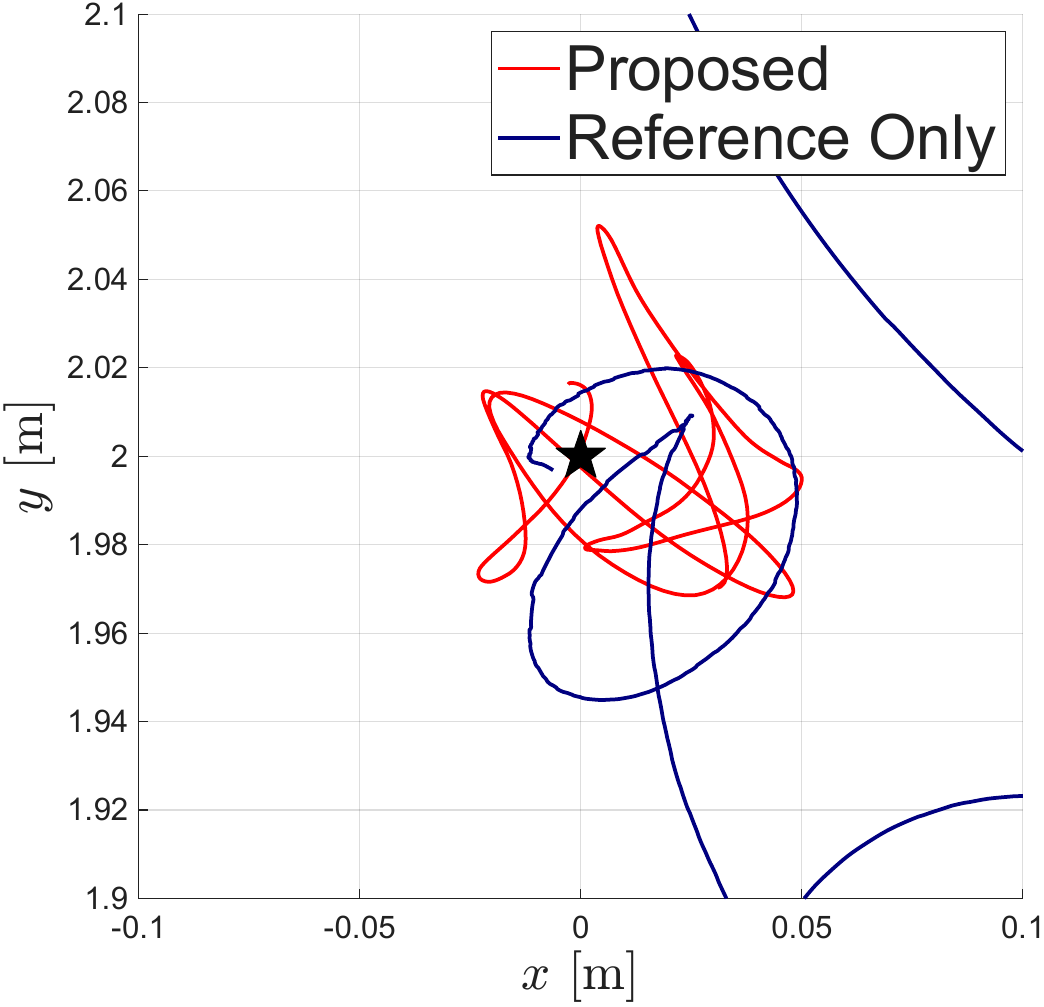}
        (d) Severe delay
    \end{minipage}
    \caption{Position trajectories under various delay levels with and without delay compensation in stabilization task of the simulation experiment. The $\bigstar$ symbols denote the stabilization target positions.}
    \label{fig:stabilization}
\end{figure*}

\begin{figure*}[t]
    \centering

    \begin{minipage}{0.24\linewidth}
        \centering
        \includegraphics[width=\linewidth]{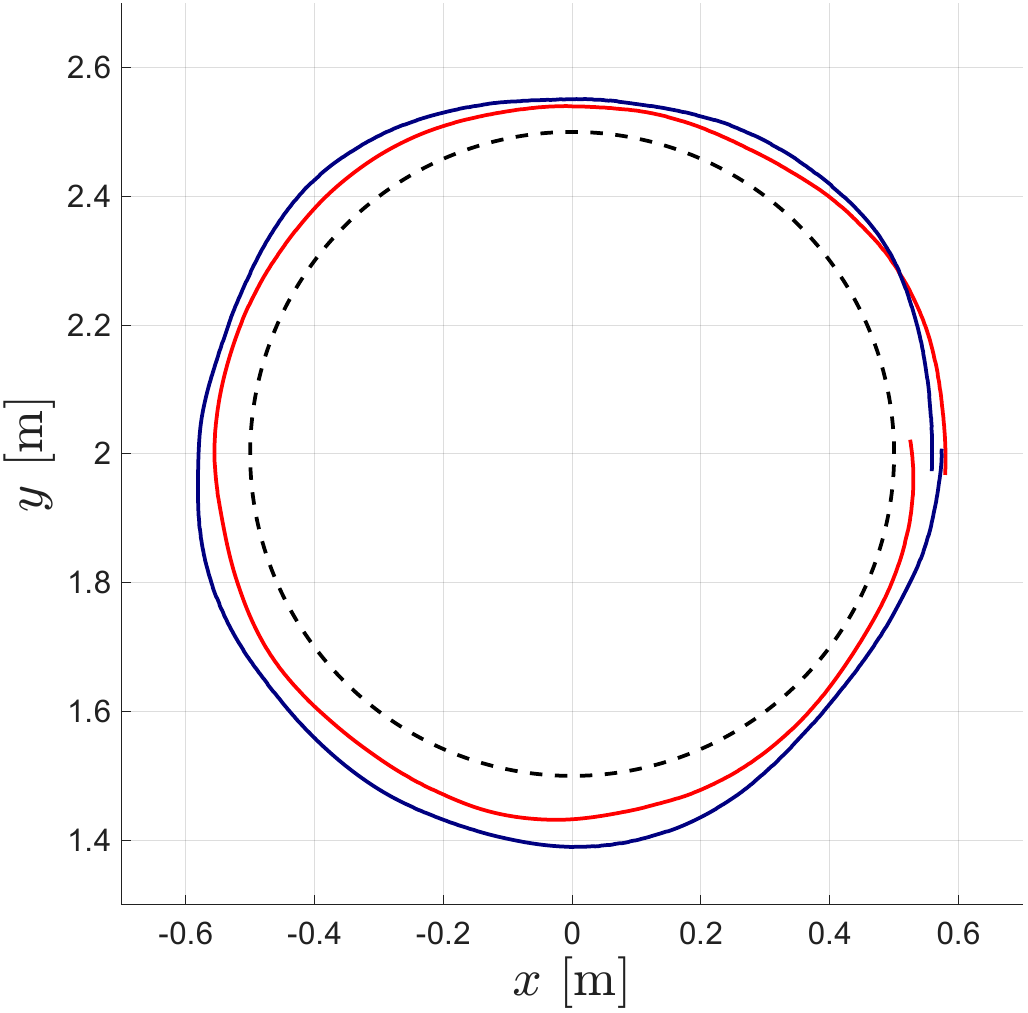}
        (a) No delay
    \end{minipage}
    \begin{minipage}{0.24\linewidth}
        \centering
        \includegraphics[width=\linewidth]{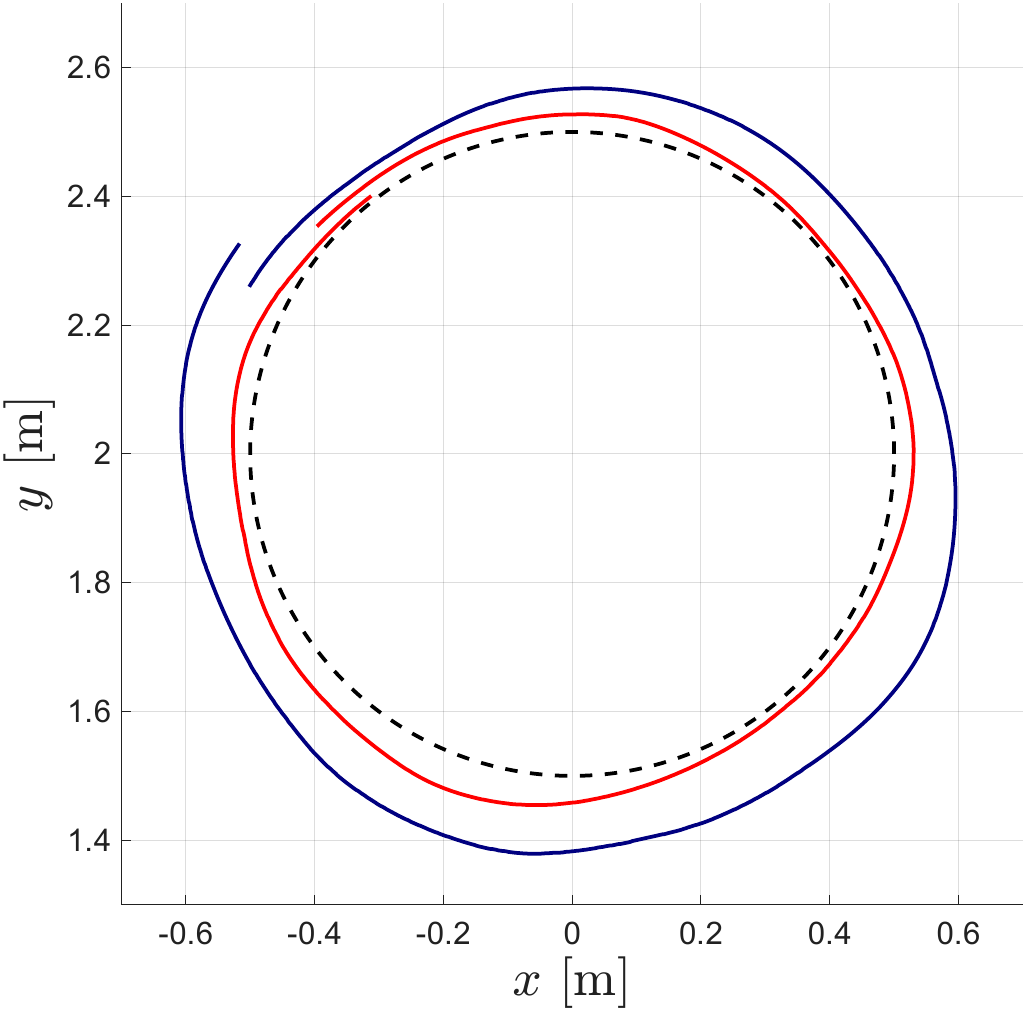}
        (b) Light delay
    \end{minipage}
    \begin{minipage}{0.24\linewidth}
        \centering
        \includegraphics[width=\linewidth]{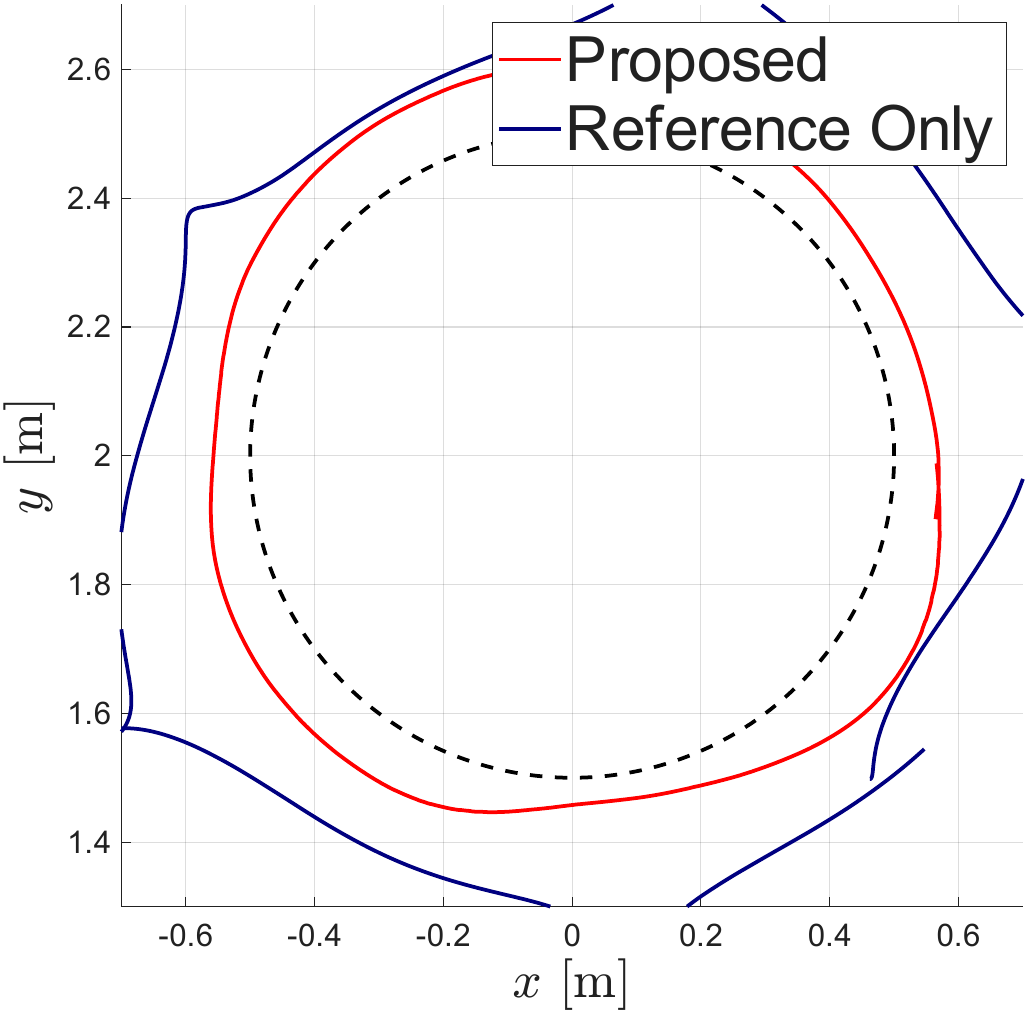}
        (c) Moderate delay
    \end{minipage}
    
    \caption{Position trajectories under various delay levels with and without delay compensation in the tracking task of the simulation experiment. The reference circular trajectory is drawn as black dotted curves.}
    \label{fig:tracking}
\end{figure*}

In the simulation experiment, we consider two tasks.
The \textit{stabilization} task is to  stabilize the robot to a fixed point. In the \textit{tracking} task, the robot is required to track a circular trajectory of 1-meter diameter, facing the center of the circle. The speed along the trajectory remains constant, and one revolution corresponds to 30 seconds.
For each task, four different delay characteristics were considered, which simulate the delay according to
\begin{equation}
    \Delta_i = s_i^u - t^u_i = \operatorname{clip}(a + b \cdot \mathcal{U}_{[0, 1]}^i, [a, b]),
\end{equation}
where $a > 0$ and $b > 0$ are parameters controlling the severity of delay, the function $\operatorname{clip}$ clips the first argument to fit in the specified interval, and $\mathcal{U}_{[0,1]}^i$-s are independent and identically distributed (i.i.d.) samples from a uniform distribution over the interval $[0,1]$. In addition to this randomness, we regard part of the messages as lost, according to a constant drop rate. The parameter values for the delay profiles considered in the simulations are summarized in Table~\ref{tab:delay parameters}, ranging from no delay to severe delay.
In all tasks, the reference input $\boldsymbol{u}_\mathrm{ref}$ is selected as a well-tuned proportional-derivative (PD) reference tracking controller. The CLF was designed using the linear quadratic regulator (LQR) technique with respect to the linearized dynamics. 

In Table~\ref{tab:simulation rmse}, we evaluate the performance of the proposed control strategy and compare with the naive approach where only the reference PD controller was used. The evaluation and comparison are done with respect to the root mean square error (RMSE) of position and yaw angles. In Fig.~\ref{fig:stabilization} and Fig.~\ref{fig:tracking}, we depict the position trajectories over thirty-second intervals. The severe delay case of the tracking task was omitted, where both controllers failed to stabilize to the reference trajectory, and thus, the RMSE metric and the trajectories lost significance.

As expected, in both controllers, the tracking error increases as the delay severity escalates. It is also a clear trend in the recorded data that the proposed method provides a more stable maneuver of the robot with significantly reduced position and orientation tracking errors, both in the stabilization and tracking tasks. It is also notable that the difference between the two controllers is not significant where delay is not severe, because the controller (\ref{eq:clf controller}) was designed to not overly interfere with the reference controller if $\boldsymbol{u}_\mathrm{ref}$ is already a \textit{sufficiently stable} input that decreases the CLF value for all predicted state in $X^{+}$. In such cases where the effect of delays is not significant, the state predictions in $X^+$ shows high accuracy, and the PD reference controller $\boldsymbol{u}_\mathrm{ref}$ is already likely to be a sufficiently stable input for all particles in $X^+$, achieving $\delta = 0$ in (\ref{eq:clf controller}).

\section{Hardware Demonstration}

For hardware demonstration, we consider a docking task in which the ATMOS robot is operated under remote closed-loop control over the internet to dock to a docking station installed on the boundary of the low-friction experiment arena, as shown in Fig.~\ref{fig:thumbnail}.

\subsection{Setup}

In the experiment setup, the ATMOS robot in Stockholm, Sweden, is controlled from the controller which is physically located in Seoul, Korea. This is approximately 7450 \si{km} away from the robot, as shown in Fig.~\ref{fig:globe}.

To convey the control and feedback messages over the network, we use FleetMQ as the middleware. 
FleetMQ is a dynamic middleware platform designed for real-time communication between distributed robotic systems operating across heterogeneous networks. FleetMQ automatically establishes the shortest peer-to-peer network route between the controller and the robot.


The experiment is conducted on a planar microgravity-emulating testbed, where the free-flying ATMOS robot moves on a flat low-friction arena. The platform floats on three passive air-bearings, rendering its in-plane dynamics effectively weightless. The robot is actuated via eight solenoid air-valves at $10$ Hz and is equipped with a rigid conic anchoring tool for docking with a mating passive docking station placed at the edge of the arena.

Robot motion is tracked using a high-frequency Qualisys motion-capture system. The pose measurements are fused onboard with IMU data, producing a state estimate transmitted to the controller.
More details are provided in \cite{roque2025towards}.


\subsection{Motion Planning}

The installed docking station and the mating connectors requires the robot to be controlled at a precision of less than a few centimeters for a successful docking. Thus, we employ the following two-step strategy to generate the docking trajectory, which is used as the reference $\boldsymbol{x}_\mathrm{ref}$ that is fed into the CLF-based controller (\ref{eq:clf controller}).
\begin{enumerate}
    \item The robot stabilizes to the \textit{parking configuration}. The parking configuration is set as directly facing the docking station in its front. In this step, the reference state $\boldsymbol{x}_\mathrm{ref}$ is the same as the parking configuration with zeros in the velocity components.
    \item When the robot approaches sufficiently close to the parking configuration, an optimal control problem is solved online to generate a feasible docking trajectory and the corresponding reference for the CLF controller.
\end{enumerate}

In this demonstration, we set the parking position to be 50 cm away from the docking station.
The optimal control problem in the second step is formulated as follows.
First, the reference yaw angle $\yaw_d$ is always set to face the docking position throughout the trajectory and is not part of the optimization. We solve the following optimization to solve for the desired positions and linear velocities as a function of time:
\begin{subequations} \label{eq:optimization1}
\begin{align}
    \min_{\pos_{(\cdot)}, \vel_{(\cdot)}, \accel_{(\cdot)}} \quad &  \sum_{k=0}^K \accel_k^\top \boldsymbol{Q_k}\accel_k 
    \label{eq:optimization_a} \\
    \text{s.t.} \quad 
    & \begin{bmatrix}
        \pos_{k+1} \\ \vel_{k+1}
    \end{bmatrix} = A\begin{bmatrix}
        \pos_{k} \\ \vel_{k}
    \end{bmatrix} 
    +B\accel_k, \label{eq:optimization_b} \\
    & \begin{bmatrix}
        \pos_{K} \\ \vel_{K}
    \end{bmatrix} = \begin{bmatrix}
        \pos_{\text{dock}} \\ \zeros{3\times 1}
    \end{bmatrix},  \label{eq:optimization_c}
    \\
    & -\accel_{\max, k} \leq \accel_k \leq \accel_{\max, k},  \label{eq:optimization_d}
\end{align}
\end{subequations}
where $k \in [0,K]$ is the time index, $\pos_k$, $\vel_k$ and $\accel_k$ are respectively the position, linear velocity, and acceleration of the trajectory, $A$ and $B$ denote the discrete-time translational dynamics (in the form of a double integrator model) of the robot. The optimization is done subject to the docking (\ref{eq:optimization_c}) and the acceleration limit (\ref{eq:optimization_d}) constraints. The acceleration limit $\accel_{\max, k} \geq 0$ is set to be a decreasing function with respect to $k$ to reduce the effect of possible deviation from the docking trajectory caused by applying high accelerations.

\subsection{Results}

\begin{figure}
    \centering
    \includegraphics[width=\linewidth, page=1]{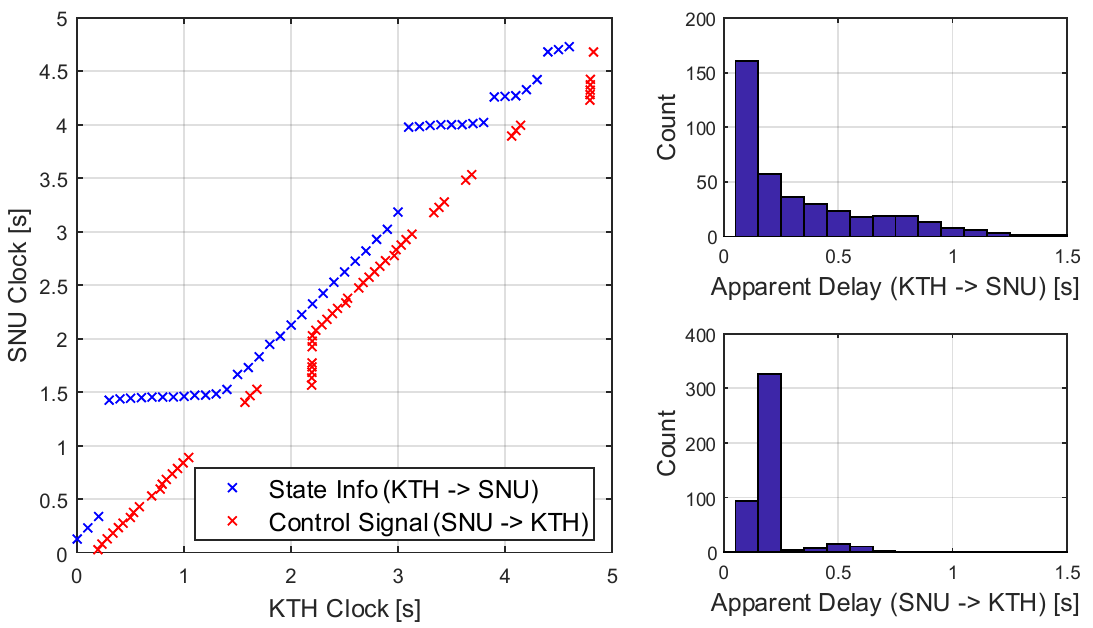}
    \caption{(Left) The message transmission and arrival times. Each point corresponds to one message, whose horizontal and vertical coordinates representing the time when it first appeared on the KTH and the SNU sides, respectively. (Right) The histograms of the apparent transmission delay. Since the two clocks cannot (and need not) be perfectly synchronized, the one-way delay is not accurately observable and the apparent delay might be biased.}
    \label{fig:experiment_delay}
\end{figure}

\begin{figure}
    \centering
    \includegraphics[width=0.75\linewidth]{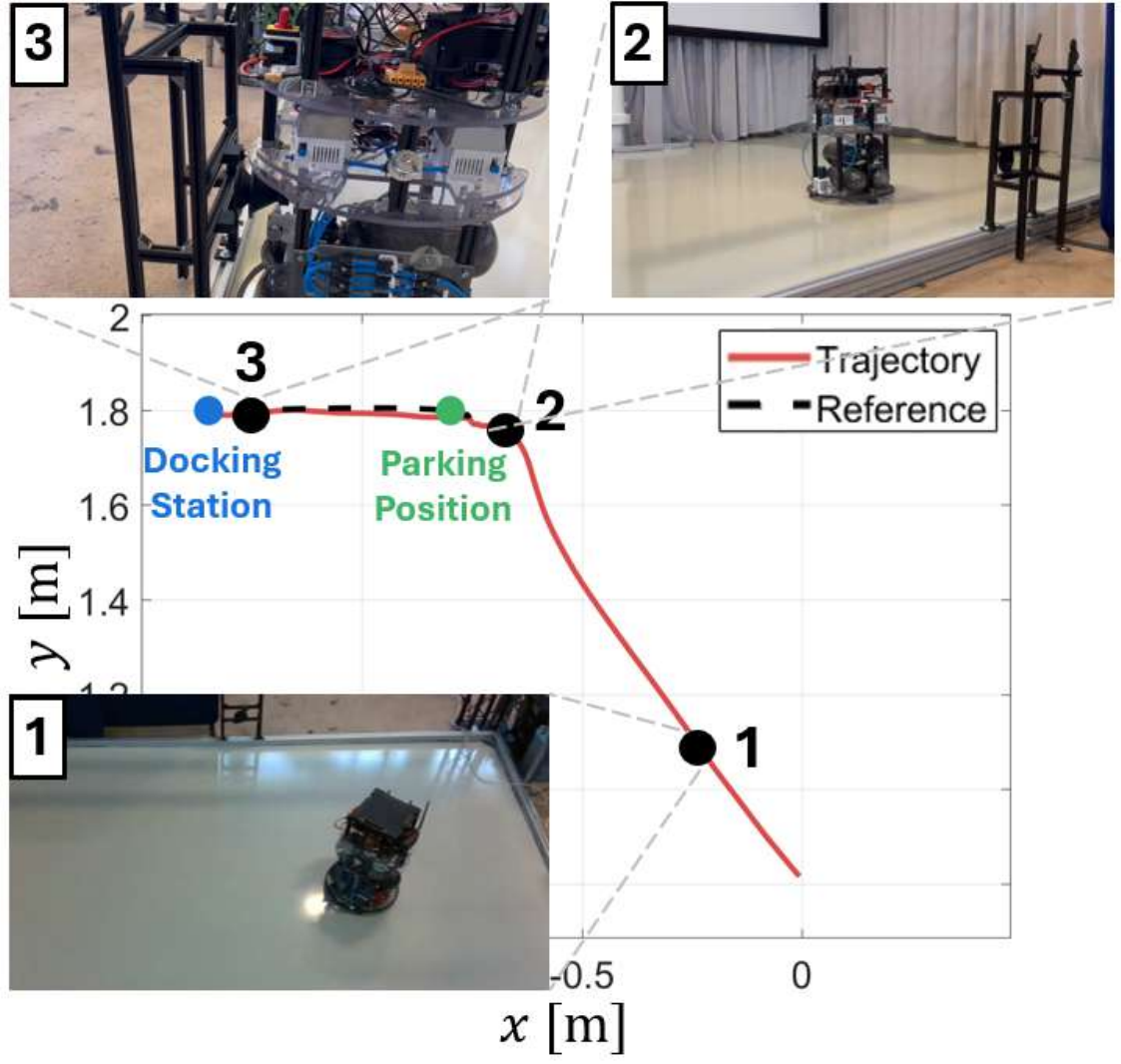}
    \caption{The position trajectory of the hardware experiment. The robot first stabilizes to the docking configuration marked green, and then the optimal trajectory computed from the optimal control problem (\ref{eq:optimization1}) is executed.}
    \label{fig:exp_figure_topdown}
\end{figure}

\begin{figure}
    \centering
    \includegraphics[width=0.90\linewidth]{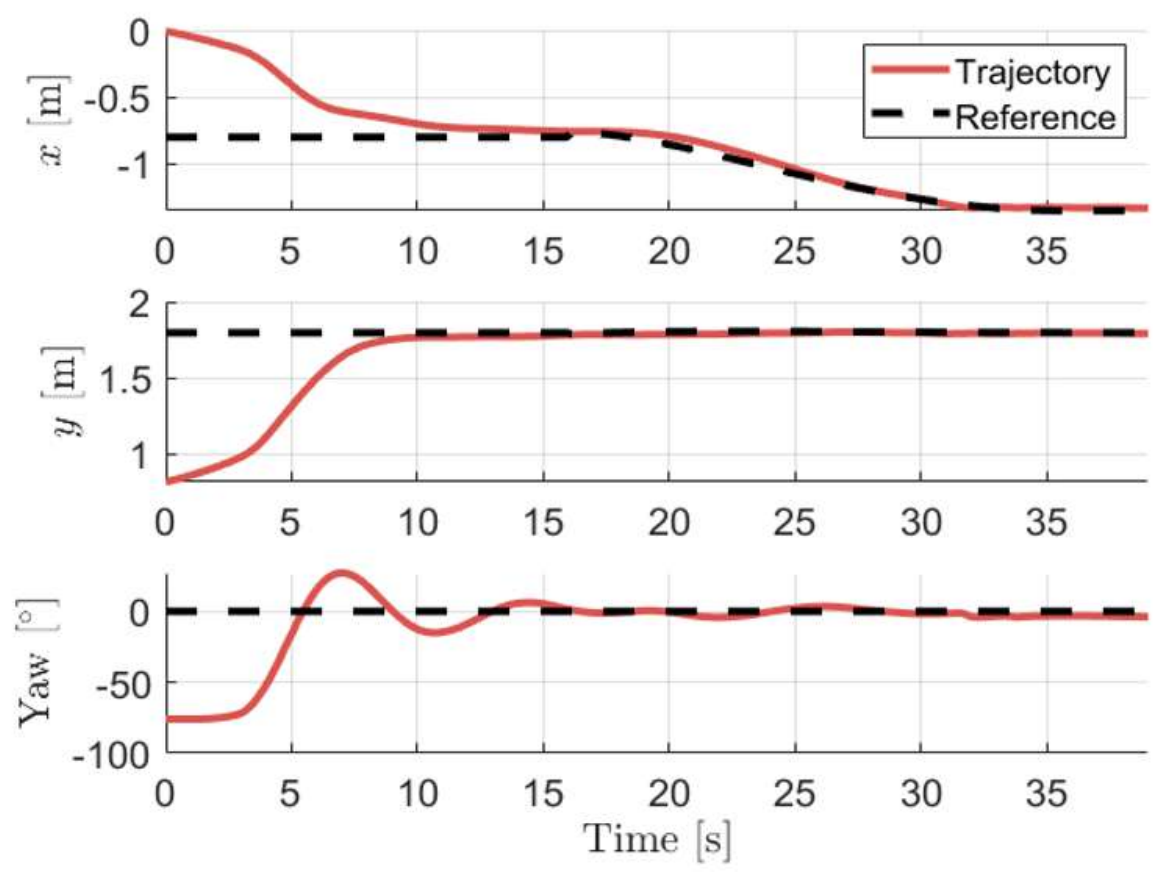}
    \caption{The configuration trajectory recorded in the hardware experiment. Even under losses of data and transmission delays, the actual robot trajectory well converged to the desired trajectory.}
    \label{fig:exp_figure_trajectory}
\end{figure}

Fig.~\ref{fig:experiment_delay} depicts the transmission delay experienced by both sides during the experiment.
The actual delay was normally constant, lying between 100 \si{ms} and 200 \si{ms} (which is about the time required to make a round-trip between the two sides in the speed of light), with intermittent blockages of up to 1 second long every 3 to 5 seconds.
Here, the source of transmission delay is not only the physical time required for the information to travel, but also the bottleneck effect due to limited network bandwidth, etc.
Indeed, this delay is highly time-correlated, and thus, the assumptions made in the belief prediction step do not hold. However, as seen in Fig.~\ref{fig:exp_figure_topdown} and Fig.~\ref{fig:exp_figure_trajectory}, the proposed control strategy showed satisfactory performance and successfully accomplished the task, proving the robustness of the approach.



\section{Conclusion}
In this work, we demonstrated a remotely controlled docking task of the ATMOS robot, a hardware platform for research in space robotics. We develop a remote control scheme that combines state prediction against the non-stationary and stochastic transmission delay with a CLF-based controller that mitigates the effects of uncertainties and possible distribution mismatches deriving from the stochastic nature of the experimental setting. Our demonstration successfully showcased that the proposed control scheme enables stable operation of the ATMOS robot over a 7450 \si{km} distance between Seoul (controller location) and Stockholm (ATMOS location).

We conclude by suggesting possible future research directions.
First, although empirical results suggest that the proposed control strategy enables stable remote control across a long distance, theoretical guarantees of performance and stability will be studied to improve the reliability and, therefore, practical applicability of the controller for space robotics tasks.
Another important future research direction is to minimize the amount of information exchange, possibly by combining event-based control methods such as \cite{gu2022event}, as the stochasticity in transmission delay often results from the saturation of network bandwidth.

\bibliography{references}             
                                                   







\end{document}

%% file: symbol_definition.tex
\newcommand{\nR}[1]{\mathbb{R}^{#1}}		

\newcommand{\zeros}[1]{\boldsymbol{0}_{#1}}

\newcommand{\state}{\boldsymbol{x}}
\newcommand{\sysinput}{\boldsymbol{u}}
\newcommand{\dstate}{\dot{\boldsymbol{x}}}
\newcommand{\f}{f}
\newcommand{\G}{G}

\newcommand{\redG}{\bar{G}}

\newcommand{\config}{\boldsymbol{q}}
\newcommand{\twist}{\dot{\boldsymbol{q}}}
\newcommand{\acceltwist}{\ddot{\boldsymbol{q}}}

\newcommand{\pos}{\boldsymbol{p}}

\newcommand{\vel}{\boldsymbol{v}}

\newcommand{\accel}{\boldsymbol{a}}

\newcommand{\yaw}{\psi}
\newcommand{\yawrate}{\omega_z}
\newcommand{\torquez}{\tau_z}
\newcommand{\inertiaz}{I_z}
\newcommand{\rotz}{R_z}

\newcommand{\force}{\boldsymbol{f}}

\newcommand{\thrustersforces}{\mathbf{v}}

\newcommand{\allocmat}{B}

\newcommand{\atmos}{ATMOS}



%% file: sections/abstract.tex
We present a demonstration showcasing the Autonomy Testbed for Multi-purpose Orbiting Systems (ATMOS), a planar spacecraft-analog robot designed for hardware-in-the-loop evaluation of guidance and control strategies in microgravity-like conditions. Using ATMOS as the physical test platform, we investigate the design, analysis, and performance evaluation of control architectures for remotely operated spacecraft under round-trip communication delays. In this work, we develop and experimentally validate a control strategy that combines state prediction and trajectory tracking control to perform a docking maneuver, accounting for time-varying random communication latency between ground operators and the ATMOS system. The demonstration includes a long-distance remote control experiment between Seoul and Stockholm, introducing realistic intercontinental delays and variability. The results highlight the capability of ATMOS to support rapid, reliable, and cost-effective testing of spacecraft teleoperation concepts, establishing a first step toward robust validation of on-orbit operations in microgravity-like environments.

%% file: sections/system_model.tex
\atmos{} \citep{roque2025towards} is a three degrees of freedom robotic platform moving quasi-frictionless on a resin plate to simulate motion in micro-gravity environments.  The robot configuration is described by $\config = (\pos, \yaw) \in \nR{2} \times (-\pi, \pi]$, where $\pos \in \nR{2}$ denotes the robot position and $\yaw \in (-\pi, \pi]$ represents the yaw angle of the robot. We indicate by $\twist = \begin{bmatrix}
    \vel^\top & \yawrate
\end{bmatrix}^\top \in \nR{3}$ the robot velocity twist, where $\vel \in \nR{2}$ is the world-frame linear velocity and $\yawrate \in \nR{}$ is the angular rate. The robot is actuated through the wrench $\sysinput = \begin{bmatrix}
    \force^\top & \torquez
\end{bmatrix}^\top \in \nR{3}$, consisting of body-frame force $\force \in \nR{2}$ and torque $\torquez \in \nR{}$. In practice, the wrench command is mapped to the robot thrusters $\thrustersforces \in \nR{8}$ through the allocation matrix $\allocmat \in \nR{3 \times 8}$ as $\sysinput = \allocmat \thrustersforces$, where $\allocmat$ has full row rank.

The full nonlinear model of the robot is written as
\begin{equation}
    \begin{aligned}
        \acceltwist
        =
        \redG(\yaw) \sysinput
    \end{aligned}
\end{equation}
with
\begin{equation}
    \redG(\yaw) = \begin{bmatrix}
            \frac{1}{m}\rotz(\yaw) & \zeros{2\times 1}
            \\
            \zeros{{1\times 2}} & \frac{1}{\inertiaz}
        \end{bmatrix}
\end{equation}
where $m$ is the mass of the robot, $\inertiaz$ is the robot inertia around the $z$-axis and $\rotz(\yaw)$ is the rotation matrix of an angle $\yaw$ about the $z$-axis. For convenience, we also define the state vector $\state = \begin{bmatrix}
    \config^\top & \twist^\top
\end{bmatrix}^\top \in \mathbb{R}^{5}\times (-\pi, \pi]$, such that the system dynamics can be written in input-affine form:
\begin{equation} \label{eq:dynamics}
    \dstate = \f(\state) + \G(\state)\sysinput
\end{equation}
where $\f(\state) = \begin{bmatrix}
    \twist^\top & \zeros{{1\times 3}}
\end{bmatrix}^\top$ and
$\G(\state) = \begin{bmatrix}
       \zeros{{3\times 3}} & \redG(\yaw)^\top
\end{bmatrix}^\top$.

%% file: sections/control_methodology.tex
In this section, we describe the control strategy to remotely operate the ATMOS robot. Namely, the controller runs on a remote computer, which sends control inputs to the robot and receives back state feedback through the communication network.
Several challenges arise from communication delays between the two sides, which we summarize as follows.

First, the robot and the controller operate on unsynchronized clocks, introducing a so-called clock bias. Secondly, the state at which a control input will be applied to the robot differs temporally from the state considered to compute such input in a feedback manner, due to transmission time delays between the robot and the controller. Moreover, the delays appear to be stochastic, some messages may be lost entirely, and the characteristics of this delay also change, albeit slowly, over time.

To address these challenges, the proposed control strategy includes a prediction step that estimates the state at which the current control inputs will be applied, representing this estimate as a distribution or belief. This predicted state is then used by a CLF-based controller, ensuring that the applied control is consistent with the estimated state. 

%% file: sections/state_prediction.tex
Suppose, at time $t^{u}_i$ on the controller clock, the controller sends a timestamped and indexed control input $\vec{u}_i$ as a packet $U_i:=(\vec{u}_i,t^u_i)$. Denote by $s^u_i$ the time on the robot clock when the $i$-th control packet arrives at the robot. 
If a packet is lost during transmission, we regard it as having arrived simultaneously with the next packet that successfully reaches the robot. For example, if the packets sent at $t^u_i$ and $t^u_{i+1}$ are both lost and the $(i+2)$-th packet is the first to arrive after that, we treat this as $s^{u}_i=s^{u}_{i+1}=s^u_{i+2}$. Additionally, if a packet $U_i$ arrives later than $U_{i+1}$ (i.e., $s^u_{i} \geq s^u_{i+1}$), it is discarded and considered lost, since it cannot overwrite a newer command.
We assume that the control packet delay $\Delta_i = s^u_i - t^u_i$, accounting for both transmission delays and clock bias, follows a quasi-stationary distribution and that delays are uncorrelated.

At the same time, the robot sends back to the controller a state packet $X_k:=(\vec{x}_k,s_{k}^x,i_{k}, \Gamma_k)$ where $\vec{x}_k = \vec{x}(s^x_k)$ is the recorded state at the time $s_{k}^x$ in which the state packet is sent, $i_k \in \mathbb{N}$ is the index of the the latest applied control input  $\vec{u}_{i_k}$ from packet $U_{i_k}$ and $\Gamma_k = (\Delta_1, \ldots \Delta_{N_{s}})$ is a continuously updated list of samples of control packet delays with $N_{s}>0$. When a state packet $X_k$ is received at the controller side, a new control packet $U_k:= (\vec{u}_k,t^u_k)$ is computed, which will be applied on the robot at the unknown time $s^u_k$ and state $\vec{x}^{+}_k:=\vec{x}(s^u_k)$ according to the controller shown later in \eqref{eq:clf controller}. An estimate  $\hat{\vec{x}}^{+}_k$ of the state $\vec{x}^{+}_k$ is obtained by simulating the state evolution, after the application of all the \textit{pending} control packets $U_{j}, \; i_{k}+1 \leq j \leq k$ (i.e., packets that were not yet applied before the state packet $X_k$ was sent). Namely, for each pending packet $U_{j}, \; i_{k}+1 \leq j \leq k$, the arrival time $\hat{s}^{u}_j$ is estimated as
\begin{equation}
    \hat{s}^{u}_j = t^u_j + \hat{\Delta}_j,\;
\end{equation}
where $\hat{\Delta}_j$ is randomly sampled from $\Gamma_k$. At this point, an estimate of the state $\hat{\vec{x}}^{+}_k$, where the currently computed packet $U_{k}$ will be received and applied from the robot, is obtained assuming the zero-order hold input signal $\vec{u}: [s^x_{k}, \hat{s}^u_{k}] \rightarrow \mathbb{R}^{m}$ is applied on the system such that, for each interval $[\hat{s}^u_j, \hat{s}^u_{j+1})$ with $i_{k}+1 \leq j <  k $, it holds
\begin{equation}
\vec{u}(t) = \vec{u}_{j} ,\; \forall t\in [\hat{s}^u_j, \hat{s}^u_{j+1}).
\end{equation}
Thus $\hat{\vec{x}}^{+}_{k}$ is predicted as 
\begin{equation}\label{eq:state prediction}
\hat{\vec{x}}^{+}_{k} = \hat{\vec{x}}(\hat{s}^{u}_{k}) = \int_{s^x_{k}}^{\hat{s}^u_{k}} \left( f(\vec{x}(t)) + G(\vec{x}(t))\vec{u}(t) \right) dt + \vec{x}_k.
\end{equation} 
Since the predicted arrival times $\hat{s}^u_j$ may not preserve their original order, we consider the packages such that $\hat{s}^u_{j} \geq \hat{s}^u_{j+1}$ as lost.

Performing the prediction in \eqref{eq:state prediction} for $N \in \mathbb{N}$ different samples of the delay profiles $\hat{s}^u_j$, we define a set $X_k^{+} = \{\hat{\vec{x}}^{+}_{k,l}, \ldots \hat{\vec{x}}^{+}_{k,N} \}$ of \textit{particles} which approximates the predicted state distribution at which the currently generated input $\vec{u}_k$ arrives at the robot.

%% file: sections/delay_compensation_control.tex
Under the given settings, it is desired to stabilize the system around a reference state $\vec{x}_\mathrm{ref}$. Let $V(\boldsymbol{x}; \boldsymbol{x}_\mathrm{ref})$ be a CLF designed to stabilize the original delay-free system (\ref{eq:dynamics}) to $\boldsymbol{x}_\mathrm{ref}$ i.e.,  ($V(\boldsymbol{x}_\mathrm{ref}; \boldsymbol{x}_\mathrm{ref})=0$ and $V(\boldsymbol{x}; \boldsymbol{x}_\mathrm{ref})>0\; \forall \vec{x} \neq \boldsymbol{x}_\mathrm{ref} $). 
If the state $\vec{x}^{+}_{k}$, in which the control input $\vec{u}_k$ will take effect, was known, then the system could be stabilized toward $\boldsymbol{x}_\mathrm{ref}$ if $\vec{u}_k$ is such that $\dot{V}(\boldsymbol{x}^{+}_k,\boldsymbol{u}_k;\boldsymbol{x}_\mathrm{ref}, \dot{\boldsymbol{x}}_\mathrm{ref}) + \gamma V(\boldsymbol{x}^{+}_k;\boldsymbol{x}_\mathrm{ref}) \leq0$, for a given $\gamma\in \mathbb{R}_{\geq 0}$ controlling the exponential convergence rate of the CLF.\footnote{The time derivative for $V(\vec{x},\vec{x}_{\mathrm{ref}})$ is given by $\dot{V}(\boldsymbol{x},\boldsymbol{u};\boldsymbol{x}_\mathrm{ref}, \dot{\boldsymbol{x}}_\mathrm{ref}) = \frac{\partial{V}}{\partial \vec{x}}(\vec{x},\vec{x}_{\mathrm{ref}})(f(\vec{x}) + G(\vec{x})\vec{u}) + \frac{\partial{V}}{\partial \vec{x}_{\mathrm{ref}}}(\vec{x},\vec{x}_{\mathrm{ref}})\dot{\vec{x}}_{\mathrm{ref}}$.}
Since we only have a set of particles $X^{+}_{k}$ to represent the true state $\boldsymbol{x}^{+}_k$, we propose to select $\vec{u}_k$ by solving
\begin{equation} \label{eq:clf controller}
\begin{aligned}
    \begin{bmatrix}
    \vec{u}^T_{k} \; \delta_k\end{bmatrix}^T = &\; \operatorname{argmin}_{\boldsymbol{u}, \delta} \quad  \lVert{\boldsymbol{u} - \boldsymbol{u}_\mathrm{ref}}\rVert^2 + p\delta \\
    \operatorname{s.t.} \; & \delta \geq 0, \\
    & \dot{V}(\hat{\boldsymbol{x}}^{+}_{k,l},\boldsymbol{u};\boldsymbol{x}_\mathrm{ref}, \dot{\boldsymbol{x}}_\mathrm{ref}) + \gamma V(\hat{\boldsymbol{x}}^{+}_{k,l};\boldsymbol{x}_\mathrm{ref}) \leq \delta, \\
    &\qquad\qquad\qquad\qquad\qquad\qquad \forall \hat{\vec{x}}_{k,l} \in X^{+}_k 
\end{aligned}
\end{equation}
where $\delta$ is a slack variable that relaxes the CLF decrease condition when strict satisfaction is not possible for all predicted states; $p \in \mathbb{R}_{\geq 0}$ is a penalty weight,
$\dot{\boldsymbol{x}}_\mathrm{ref}$ is the state velocity of the reference trajectory, $\boldsymbol{u}_\mathrm{ref}$ is the reference input typically deriving from a stabilizing controller or a manual control signal from an operator.

The optimization-based controller (\ref{eq:clf controller}) tries to search for the input $\boldsymbol{u}_k$ such that 1) $\boldsymbol{u}_k$  minimally deviates from the reference $\boldsymbol{u}_\mathrm{ref}$ and 2) $\boldsymbol{u}_k$ enforces an exponential decrease of the CLF (or a minimal increase thereof), for all possible predicted states $\hat{\vec{x}}^{+}_{k,l}\in X_k^{+}$. 
These two potentially conflicting objectives are balanced via the positive tunable parameter $p$, where a bigger $p$ implies a priority on the satisfaction of the CLF constraint, resulting in a more robust but conservative behavior.

Considering the input-affine dynamics in (\ref{eq:dynamics}), the optimization in \eqref{eq:clf controller} is a standard quadratic program with linear constraints, which can be solved efficiently, and is always feasible. While formal guarantees on the stability of the system under  (\ref{eq:clf controller}) with stochastic delays are difficult and omitted here, the fact that the CLF constraint $\dot V + \gamma V \le \delta$ is enforced for all predicted states $\hat{\state}^{+}_{k,l} \in X_k^+$ naturally yields a worst-case-aware controller that provides some robustness toward the assumption considered over the time delay distribution. This robustness property was indeed confirmed in simulation and hardware experiments.